\documentclass[sigconf]{acmart}
\usepackage{amsmath}

\AtBeginDocument{%
  }

\setcopyright{acmcopyright}
\copyrightyear{2023}
\acmYear{2023}
\acmDOI{XXXXXXX.XXXXXXX}

\acmConference[]{}
\acmBooktitle{}
\acmPrice{15.00}
\acmISBN{978-1-4503-XXXX-X/18/06}
\renewcommand\footnotetextcopyrightpermission[1]{}
\acmSubmissionID{1743}

\begin{document}

\title{VideoControlNet: A Motion-Guided Video-to-Video Translation Framework by Using Diffusion Model with ControlNet}


\author{Zhihao Hu}
\affiliation{%
  \institution{Beihang University, China}
  \country{}
}

\author{Dong Xu}
\affiliation{
  \institution{University of Hong Kong, China}
  \country{}
}







\begin{abstract}
  Recently, diffusion models like StableDiffusion have achieved impressive image generation results. However, the generation process of such diffusion models is uncontrollable, which makes it hard to generate videos with continuous and consistent content. In this work, by using the diffusion model with ControlNet, we proposed a new motion-guided video-to-video translation framework called VideoControlNet to generate various videos based on the given prompts and the condition from the input video. Inspired by the video codecs that use motion information for reducing temporal redundancy, our framework uses motion information to prevent the regeneration of the redundant areas for content consistency. Specifically, we generate the first frame (\textit{i.e.,} the I-frame) by using the diffusion model with ControlNet. Then we generate other key frames (\textit{i.e.,} the P-frame) based on the previous I/P-frame by using our newly proposed motion-guided P-frame generation (MgPG) method, in which the P-frames are generated based on the motion information and the occlusion areas are inpainted by using the diffusion model. Finally, the rest frames (\textit{i.e.,} the B-frame) are generated by using our motion-guided B-frame interpolation (MgBI) module. Our experiments demonstrate that our proposed VideoControlNet inherits the generation capability of the pre-trained large diffusion model and extends the image diffusion model to the video diffusion model by using motion information. More results are provided at our \href{https://vcg-aigc.github.io/}{\textcolor{magenta}{project page}}.
\end{abstract}





\begin{teaserfigure}
  \centering    
  \includegraphics[width=0.99\textwidth]{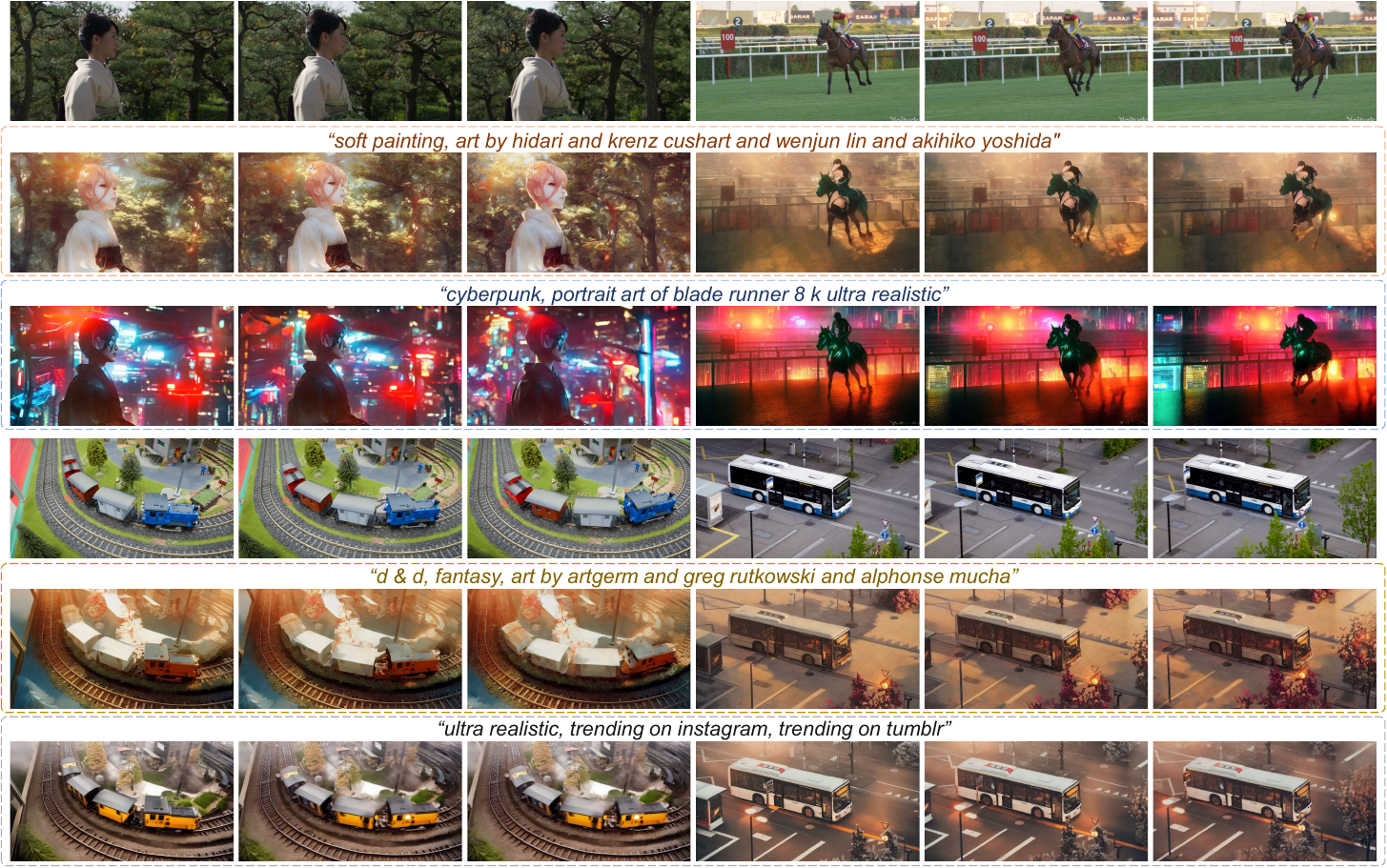}
  \caption{Various generation results when using different input videos and different prompts. The first and the fourth rows are the input videos and the other rows contain the generation results when using the corresponding prompts.}
  \Description{x.}
  \label{fig:performance}
\end{teaserfigure}


\maketitle

\section{Introduction}

Video generation is an essential task in computer vision. Most previous works~\cite{wang2018video,wang2019few} leverage the generative adversarial networks (GAN)~\cite{goodfellow2020generative} for generating continuous and content-consistent videos. Recently, diffusion models~\cite{ho2020denoising,rombach2022high} have attracted increasing attention, which has shown strong generation capability over generative adversarial networks. More recently, the StableDiffusion model is released and achieves state-of-the-art generation performance, which is trained on the large-scale text-image dataset and thus can generate various types of images based on the given text prompt. Although image diffusion models have achieved promising results these days, recently released video diffusion models~\cite{singer2022make,blattmann2023align,ho2022video} fail to generate continuous and content-consistent videos with high quality.

The main reason leading to the failure of video diffusion models is the uncontrollable diffusion process. Given the input text prompt, the diffusion process is uncontrollable and various types of images may be generated. Fortunately, ControlNet~\cite{zhang2023adding} is recently proposed for controlling the generation process of diffusion model based on different conditions (\textit{e.g.}, canny map, depth map or segmentation map, etc.). Therefore, it is intuitive to generate the output video based on the condition from the given input video. However, when the output video is directly generated frame-by-frame, it is still hard to promise the content consistency of the neighbor frames. One possible explanation is that one condition still corresponds to various output content and thus also results in inconsistent content between the independently generated frames.

Inspired by the video coding process~\cite{lu2019dvc,hu2021fvc,hu2022fvc} that adopts the motion information for reducing temporal redundancy, we propose a new motion-guided video-to-video translation framework called VideoControlNet by using the diffusion model with ControlNet, in which the motion information is adopted for content-consistency and the diffusion-model-based inpainting is used for covering the residual information. Therefore, by using the video coding paradigm that uses motion information for reducing redundancy, our method prevents the regeneration of the redundant areas based on the motion information and thus we can keep better content consistency. Specifically, we set the first frame as the I-frame and divide the following frames into different groups of pictures (GoP), in which the last frame of different GoPs is set as the key frame (\textit{i.e.,} P-frame) and other frames are set as B-frames. We first generate the I-frame independently by directly using the diffusion model with ControlNet, in which the condition is extracted from the I-frame of the input video and thus the output I-frame has the same content structure as the input I-frame. Then we generate the rest P-frames by using our newly proposed motion-guided P-frame generation (MgPG) module, in which the motion information is used for motion compensation of the redundant areas and the diffusion-model-based inpainting is performed for the generation of the newly occurred areas. Finally, the B-frames are generated based on our motion-guided B-frame interpolation (MgBI) module. Our proposed framework follows the paradigm of the B-frame-based video decoding process and inherits the generation capability of the image diffusion model, which makes it able to generate high-quality videos with continuous and consistent content. Therefore, guided by the motion information of the input video, our framework can generate different videos with different styles or contents based on different given text prompts.

To demonstrate the effectiveness of our proposed framework, we perform experiments based on the current most well-known StableDiffusion model, which is trained on a large-scale text-image dataset and achieves state-of-the-art generation performance for image diffusion. The generation results of our VideoControlNet based on the StableDiffusion Model with ControlNet are provided in Figure~\ref{fig:performance}. For example, for the first video of a woman walking in the forest, our method can generate the output video that has the same content as the input video and different styles (\textit{e.g.,} the soft-painting style of the artwork or the cyberpunk style of realistic photo). The results demonstrate that our method can keep the content consistent and generate various videos based on different text prompts, which inherit the spirit of the StableDiffusion Model and our VideoControlNet framework makes it able to extend the StableDiffusion model to the video diffusion model.

Our contributions are summarized as follows,
\begin{itemize}
    \item We proposed a new motion-guided video-to-video translation framework called VideoControlNet by using the diffusion model with ControlNet following the paradigm of video coding. 
    \item To generate the P-frames based on the given I-frame, we propose the motion-guided P-frame generation (MgPG) module, in which the motion information is extracted from the input video for keeping the content consistency, and the residual areas are generated by diffusion-model-based inpainting.
    \item We also propose the motion-guided B-frame interpolation (MgBI) module for generating the rest B-frames based on the reference I/P-frame.
    \item Experimental results demonstrate that our method inherits the generation capability of the pre-trained large diffusion model (\textit{i.e.,} StableDiffusion) and is able to translate the input video into diverse videos with different styles or contents.
\end{itemize}

\section{Related Works}

\begin{figure*}[t]
  \centering
  \includegraphics[width=\linewidth]{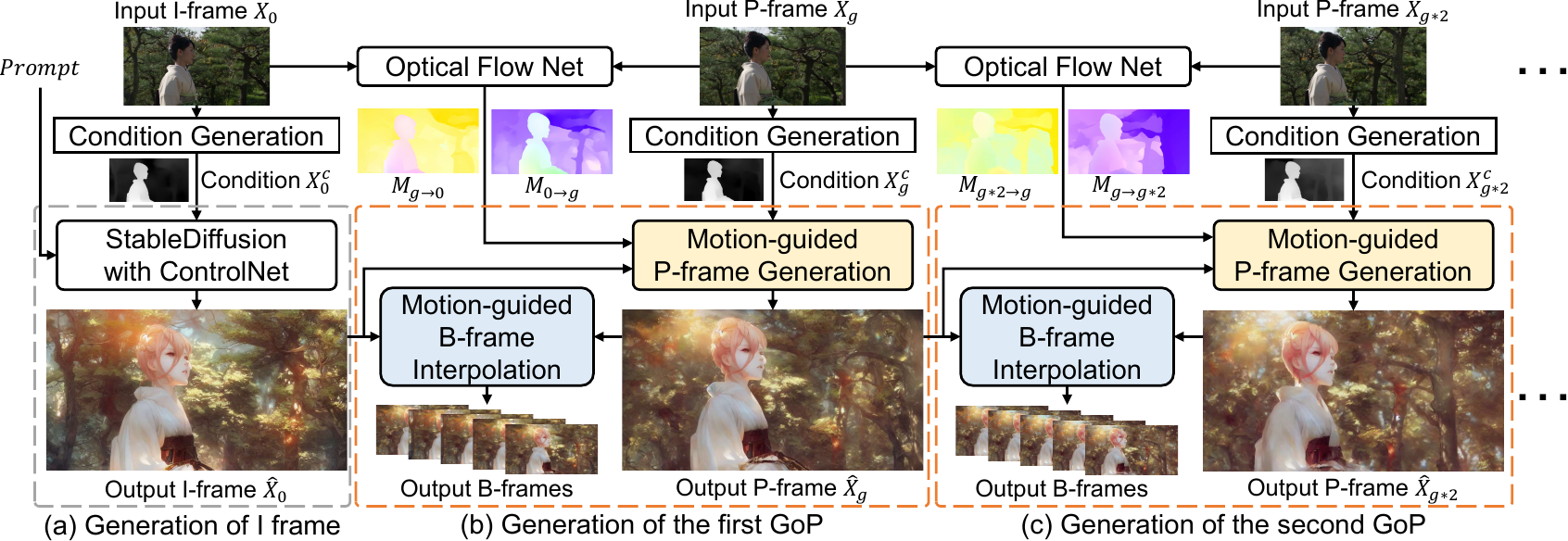}
      \caption{Overview of our proposed motion-guided video-to-video translation framework. (a) The generation process of I-frame: Taking the first input frame $X_0$ as the I-frame, we first generate the condition image $X^c_0$. Then we use the StableDiffusion with ControlNet to generate the output I-frame $\hat{X}_0$, which has the same content structure as the input I-frame $X_0$. (b) The generation process of the first Group of Pictures (GoP): We take the output I-frame as the reference frame and perform the motion-guided P-frame generation (MgPG) to generate the output P-frame $\hat{X}_{g}$. After that, the rest B-frames in the current GoP are generated by using our newly proposed motion-guided B-frame interpolation (MgBI) module. (c) The generation processes of the rest GoPs are the same as the generation process of the first GoP.}
  \label{fig:overview}
  \Description{x2.}
  \end{figure*}
\subsection{Diffusion Model}

\textbf{Image Generation.}
Diffusion models~\cite{ho2020denoising,song2020score,song2020denoising,rombach2022high} perturb data during the forward process and recover the data during the inversion process, which achieves the state-of-the-art image generation performance. Although the generation speed of the first few works~\cite{ho2020denoising,song2020denoising} are extremely slow due to their large number of diffusion steps, some works~\cite{ludpm,lu2022dpm} study to accelerate the generation speed by introducing new sampling strategies. Recently, LatentDiffusion~\cite{rombach2022high} introduced VQ-VAE~\cite{van2017neural} to diffusion models and performs the time-consuming diffusion process in latent space, which is further extended to StableDiffusion by training on large-scale text-image dataset.

\vspace{1mm}
\noindent\textbf{Control of Diffusion Model.}
Most recent state-of-the-art image diffusion models~\cite{nichol2022glide,kim2022diffusionclip,avrahami2022blended} are guided by the control of the text information, which can be achieved by extracting CLIP features~\cite{ramesh2022hierarchical} from the text and then concatenate the CLIP feature during the diffusion steps or use cross-attention module. SDEdit~\cite{meng2021sdedit} achieved controllable image editing by adding noise to the given stroke without extra training steps for diffusion models. EGSDE~\cite{zhao2022egsde} adopted the energy function to control the generation during the denoising process of a pre-trained SDE. Recently, ControlNet~\cite{zhang2023adding} is proposed for adding extra conditions (\textit{e.g.}, canny map, depth map, segmentation map, etc.) to the pre-trained diffusion models, which makes it able to control the content structure of the generated results without sacrificing the generation ability.

\vspace{1mm}
\noindent\textbf{Video Generation.}
With the success of text-to-image generation, a number of works~\cite{ho2022video,singer2022make,zhou2022magicvideo,ruan2022mm,esser2023structure,liu2023video,wu2022tune,ni2023conditional,blattmann2023align} are proposed for generating videos based on image diffusion models. Some of these works~\cite{ho2022video,singer2022make,blattmann2023align} achieved the video diffusion model by changing the 2D Unet from the image diffusion models to 3D Unet and then training the 3D Unet structure on video datasets. Imagen Video~\cite{ho2022imagen} proposed a cascaded video generation framework by performing temporal super-resolutions and spatial super-resolutions on the initially generated low frame-rate and low-resolution videos. Tune-A-Video~\cite{wu2022tune} proposed a one-shot tuning strategy to enhance temporal consistency. MMDiffusion~\cite{ruan2022mm} achieved simultaneous audio and video generation. \cite{ni2023conditional} proposed the latent flow diffusion model to generate video from a single reference image, while they can only achieve low-resolution generation on specific datasets. VideoLDM~\cite{blattmann2023align} achieves the state-of-the-art text-to-video generation results by adding additional temporal layers to the pre-trained latent diffusion model~\cite{rombach2022high} and uses a cascaded framework for generating high-resolution and high frame-rate videos. However, the state-of-the-art methods do not leverage the motion information for preventing the regeneration of redundant areas, which still leads to content inconsistency.

\subsection{Video-to-Video Translation}
Image-to-image translation algorithms like pix2pix~\cite{isola2017image} are able to achieve video generation by processing the input video frame by frame, which cannot consider the temporal consistency of neighbor frames. Early video-to-video translation methods~\cite{wang2018video,wang2019few} are proposed based on the generative adversarial networks (GAN)~\cite{goodfellow2020generative}. However, such GAN-based networks can only synthesize videos based on specific training data. Recently, inspired by the strong generation capability of diffusion models, some methods also adopt diffusion models for video editing~\cite{liu2023video} and video-to-video translation~\cite{esser2023structure}. Gen-1~\cite{esser2023structure} proposed the structure-guided video generation algorithm by introducing the depth map as the condition, which needs to retrain the diffusion model on video data. Different from Gen-1, our framework is built upon the pre-trained large image diffusion model (\textit{i.e.,} StableDiffusion) and inherits the strong generation capability of StableDiffusion. Additionally, our framework follows the paradigm of the video coding~\cite{lu2019dvc,hu2021fvc,hu2023complexity} framework and prevents regeneration of the redundant areas by using motion information from the input video, which can better keep the content consistent.

\begin{figure*}[t]
  \centering
  \includegraphics[width=0.75\linewidth]{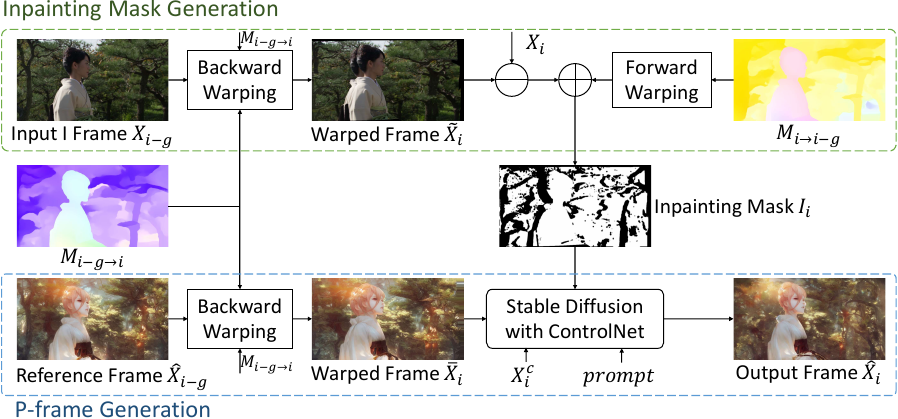}
      \caption{Details of our Motion-guided P-frame Generation (MgPG) module. In the inpainting mask generation module (\textit{i.e.,} the upper green block), we first warp the input frame $X_{i-g}$ by using the estimated optical flow $M_{i-g \rightarrow i}$ and then calculated the residual information $R_{i}$ between the warped frame $\tilde{X}_i$ and the input frame $X_{i}$. We also generate an occlusion map $O_{i}$ by using forward warping based on the optical flow $M_{i \rightarrow i-g}$. Based on the residual information $R_{i}$ and the occlusion map $O_{i}$, we generate the inpainting mask $I_{i}$. Then we perform the P-frame generation in the lower blue block by first warping the output reference frame $\hat{X}_i$ based on the optical flow $M_{i-g \rightarrow i}$ to generate the wrapped frame $\bar{X}_{i}$. Based on the inpainting mask $I_i$, the condition $X_i^c$ and the given prompt, we inpaint the uncertain areas by using the StableDiffusion with ControlNet and generate the final output P-frame $\hat{X}_i$.}
  \label{fig:mgpg}
  \Description{mgpg.}
\end{figure*}

\subsection{Optical Flow Estimation}
Our proposed VideoControlNet relies on the pre-trained optical flow estimation network for modulating the motion of the video. Therefore, it is necessary to select an effective optical flow estimation network for motion estimation. SpyNet~\cite{ranjan2017optical} is widely used in learning-based video coding frameworks like DVC~\cite{lu2019dvc}. Although such optical flow estimation networks can achieve promising results on specific datasets, they are not general for estimating the optical flow on different types of videos. Recently, FlowFormer~\cite{huang2022flowformer} is proposed and some models are trained on all the existing optical flow datasets, which is more general. To this end, we directly use the FlowFormer as our optical flow network. In the future, our method can be further enhanced by using more effective optical flow estimation networks.

\section{Method}

\subsection{Overview}

In this work, we propose a new video-to-video translation framework called VideoControlNet to translate the input video to the output video based on the given prompt. The problem formulation and the overall pipeline are provided as follows.

\noindent\textbf{Problem Formulation.} 
Given the input video $X = \{X_0, X_1, ..., X_n\}$, in which $X_i$ denotes the input frame at the timestep $i$. We denote the first input frame $X_0$ as the I frame. Then we divide the rest frames into different groups of pictures (GoP), in which we set the GoP size as $g$. We also set the last frame of each GoP as P-frame, and the other frames are denoted as B-frames. Our goal is to generate output video $\hat{X} = \{\hat{X}_0, \hat{X}_1, ... , \hat{X_n}\}$ based on the given text prompt. The corresponding frames of the output video are also called I-frames, P-frames and B-frames.

\noindent\textbf{Generation of I-frame.} 
The generation process of the I-frame is provided in Figure~\ref{fig:overview}(a). We first generate the I-frame $\hat{X}_0$ by directly performing the pre-trained StableDiffusion model with ControlNet, in which the condition $X_0^c$ (\textit{e.g.,} canny map or depth map) is extracted from the input I-frame $X_0$. As the ControlNet is able to control the content structure based on the given condition $X_0^c$, The generated I frame $\hat{X}_0$ has the same content structure as the input frame $X_0$.

\noindent\textbf{Generation of the first GoP.} 
As shown in Figure~\ref{fig:overview}(b), taking the output I-frame $\hat{X}_0$ as the reference frame, we generate the output P-frame $\hat{X}_g$ by using our newly proposed motion-guided P-frame generation (MgPG) module based on the optical flow information extracted from the optical flow net~\cite{huang2022flowformer} and the condition image $X_g^c$. After that, our motion-guided B-frame Interpolation (MgBI) module takes the output I-frame $\hat{X}_0$ and the output P-frame $\hat{X}_g$ as the reference frames and generates the output B frames in this GoP.

\noindent\textbf{Generation of the rest GoPs.} 
The generation processes of the rest GoPs are similar to the generation process of the first GoP, in which the only difference is that we use the output P-frame in the previous GoP to replace the output I-frame as the reference frame for the MgPG module and MgBI module.

\subsection{Motion-guided P-frame Generation}

As the generation process of the diffusion model is unstable, which leads to content inconsistency when generating the output video frame-by-frame. Therefore, we propose the motion-guided P-frame generation (MgPG) method, which leverages the motion information from the input video to keep content consistent by preventing the regeneration of the redundant areas. Additionally, for the areas that do not appear in the previous frames (\textit{e.g.,} the occurrence of the occlusion areas), we propose the inpainting mask generation module to generate the inpainting mask and then perform inpainting based on StableDiffusion with ControlNet.

As shown in Figure~\ref{fig:mgpg}, to generate the output P-frame $\hat{X}_i$ in the P-frame generation block, we first take the previously generated I/P-frame $\hat{X}_{i-g}$ as the reference frame and perform the backward warping operation based on the motion information (\textit{i.e.}, optical flow) $M_{i-g \rightarrow i}$ to generate the warped frame $\bar{X}_i$. It is observed that the warped frame $\bar{X}_i$ is not perfect due to the occurrence of the occlusion areas (\textit{e.g.}, at the left side of the woman and the right border in the warped frame $\bar{X}_i$ in Figure~\ref{fig:mgpg}). In the video coding task, the residual information will be added in such areas, while it is hard to generate the residual information for our generated P frame. Fortunately, the diffusion model can achieve image inpainting based on the given inpainting mask. Therefore, the key point is to generate the inpainting mask $I_i$.

\noindent\textbf{Inpainting Mask Generation.}

The inpainting mask generation process is shown in the upper green block of Figure~\ref{fig:mgpg}. Similar to the P-frame generation process, we first warp the input frame $X_{i-g}$ based on the motion information $M_{i-g \rightarrow i}$. Then the residual information can be obtained by calculating $R_i=(X_i - \tilde{X}_i)^2$. Therefore, it is intuitive to inpaint the areas with large residuals. However, only using the residual information to generate the inpainting mask is not reliable. The reason is that the RGB values in some occlusion areas are not changed too much, which makes it hard to find out all the inpainting areas by only using the residual map $R_i$ (more discussions will be provided in Section~\ref{subsec:modelanalysis}). Therefore, we additionally use the forward optical flow to find out the areas that do not appear in the reference frame. Specifically, we first estimate the inverse optical flow $M_{i \rightarrow i-g}$ and then perform the forward warping operation based on the map with values of all ones, which is formulated as follows,
\begin{equation}
    O_i = ForwardWarp(Ones, M_{i \rightarrow i-g})
\label{eq:warp}
\end{equation}
in which $Ones$ denotes the map with values of all ones. In the generated occlusion map $O_i$, areas with zeros denote they do not occur in the reference frame, which should be inpaint. After that, we generate the inpainting mask $I_i$ by considering both residual information $R_i$ and the occlusion map $O_i$, which is formulated as follows,
\begin{align}
I_{i,k} =    \begin{cases} 
                    1 & \text{if $O_{i,k} - \alpha R_{i,k} > threshold$} \\
                    0 & \text{otherwise} 
                \end{cases}
\end{align}
in which $I_{i,k}$, $O_{i,k}$, $R_{i,k}$ denotes their corresponding values at the spatial location $k$. $\alpha$ and $threshold$ are hyper-parameters. The areas with value zeros in $I_i$ will be inpaint in the diffusion model.

Finally, guided by the inpainting mask $I_i$, we inpaint the newly occurred areas in the warped frame $\bar{X}_i$ by using StableDiffusion with ControlNet and generate the output P-frame $\hat{X}_i$.

\begin{figure}[t]
\centering
\includegraphics[width=\linewidth]{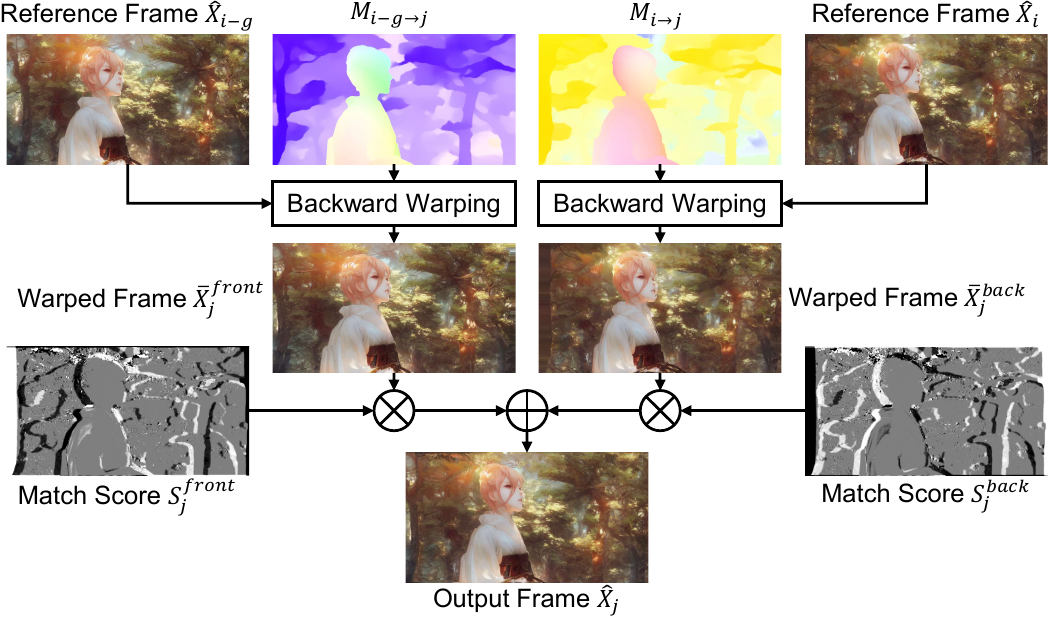}
    \caption{Details of our Motion-guided B-frame Interpolation (MgBI) module. Given two reference I/P frames $\hat{X}_{i-g}$ and $\hat{X}_i$, we aims to generate the output frame $\hat{X}_j$. We first estimate the motion information $M_{i-g \rightarrow j}$ and $M_{i \rightarrow j}$ from the corresponding input frames. Then we perform the backward warping operation to generate the warped frames $\bar{X}_j^{front}$ and $\bar{X}_j^{back}$ based on the corresponding reference frames and motion information. Based on the match score $S_j^{front}$ and $S_j^{back}$, we generate the output B-frame $\hat{X}_j$.}
\label{fig:mgbi}
\Description{mgbi.}
\end{figure}

\begin{figure*}[t]
  \centering
  \includegraphics[width=\linewidth]{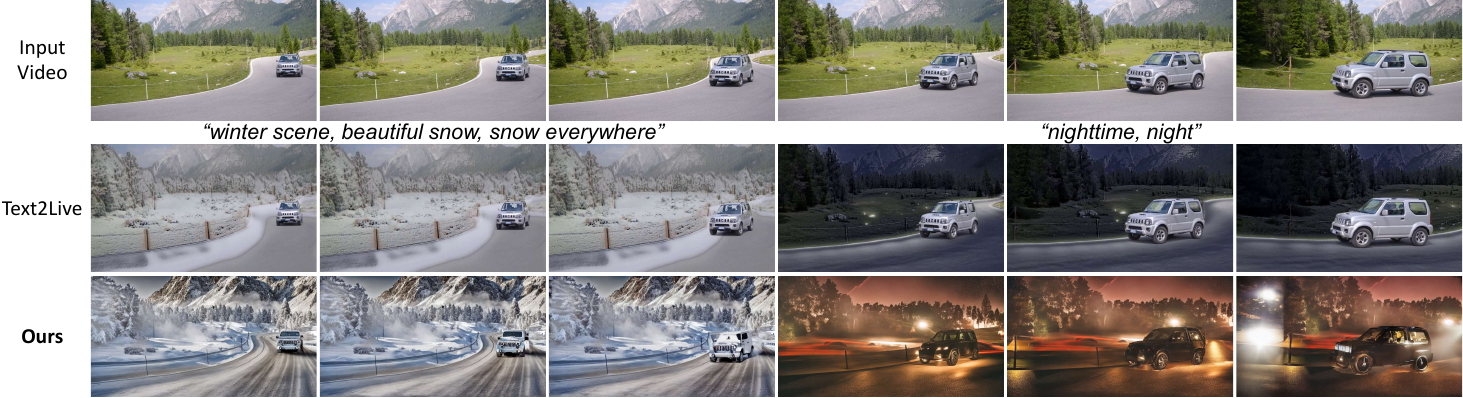}
      \caption{Generation results of Text2LIVE~\cite{bar2022text2live} and our method when using the input video at the first row and the corresponding prompts.}
  \label{fig:results}
  \Description{results.}
  \end{figure*}

\subsection{Motion-guided B-frame Interpolation}
Based on the generated output I-frame and P-frames, we generate the rest B-frames by using our motion-guided B-frame interpolation (MgBI) module. In the video coding task, the B-frame always needs less residual information than the P-frames. Therefore, in our MgBI method, we directly interpolate the B-frames based on the reference I/P frames and the motion information extracted from the input frames without using the time-consuming diffusion model.

The detailed generation process is shown in Figure~\ref{fig:mgbi}. When generating the output frame $\hat{X}_j$, we first take the two nearest I/P frames $\hat{X}_{i-g}$ and $\hat{X}_i$ as the reference frames. Then we generate the corresponding motion information $M_{i-g \rightarrow j}$ and $M_{i \rightarrow j}$ by using the optical flow net~\cite{huang2022flowformer}. After that, we perform the backward warping operation to generate the warped frames $\bar{X}_j^{front}$ and $\bar{X}_j^{back}$, in which some areas in the warped frames are inaccurate due to the occlusion areas. Considering the occlusion areas from one reference frame always occur in another reference frame, we simply generate the match score of each warped frame to produce the final output frame $\hat{X}_j$.

The match score generation process is similar to the inpainting mask generation process that also uses both residual information and forward warping operation to generate the inpainting mask. Take the match score calculation process of the warped frame $\bar{X}_j^{front}$ as an example, we first perform the backward warping operation based on the input frame $X_{i-g}$ and therefore the residual $R_j^{front}$ can be calculated. Similar to Eq.~\ref{eq:warp}, we use the motion information $M_{i-g \rightarrow j}$ to perform the forward warping operation based on the map with values of ones and thus we generate the occlusion map $O_j^{front}$. The residual map $R_j^{back}$ and the occlusion map $O_j^{back}$ are also generated in the same process. Then the intermediate score $\bar{S}_j^{front}$ and $\bar{S}_j^{back}$ are calculated as follows,

\begin{align}
    \bar{S}_{j,k}^{front} = O_{j,k}^{front} - \beta R_j^{front} \\
    \bar{S}_{j,k}^{back} = O_{j,k}^{back} - \beta R_j^{back}
\end{align}

And we use the softmax operation with temperature to generate the final match score $S_j^{front}$,
\begin{equation}
    S_{j,k}^{front} = \frac{exp(\bar{S}_{j,k}^{front})/\tau}{exp(\bar{S}_{j,k}^{front}/\tau) + exp(\bar{S}_{j,k}^{back}/\tau)}
\end{equation}
and the match score $S_j^{back}$ can be calculated by $S_j^{back} = 1 - S_j^{front}$, in which the temperature $\tau$ is the hyper-parameter.

Finally, the output B-frame $\hat{X}_j$ is calculated by adding up the weighted warped frame, which is also formulated as follows
\begin{equation}
    \hat{X}_j = S_j^{front} \times \bar{X}_j^{front} + S_j^{back} \times \bar{X}_j^{back}
\end{equation}

Therefore, our motion-guided B-frame generation module is able to generate the B-frames efficiently and effectively without using the time-consuming StableDiffusion.

\section{Experiments}

\subsection{Experimental Setup}
\textbf{Datasets.}
Due to the strong generation capability of the large StableDiffusion Model, our VideoControlNet is also general and can be applied to any type of input video. Therefore, we evaluate our method on various video datasets including the video coding datasets HEVC dataset~\cite{sullivan2012overview}, UVG dataset~\cite{UVGdataset}, and MCL-JCV datasets~\cite{wang2016mcl}. Following the previous video translation works~\cite{bar2022text2live}, we evaluate our VideoControlNet framework on the DAVIS dataset~\cite{pont20172017}, which also contains various types of videos.

\vspace{1mm}
\noindent\textbf{Implementation Details.}
In this work, we use the pre-trained StableDiffusion model~\cite{rombach2022high} in version 1.5 with the ControlNet~\cite{zhang2023adding} for I-frame generation and our motion-guided P-frame generation (MgPG) module. Specifically, we use the released models from the official GitHub repository of ControlNet, in which we use the depth map and canny map condition model. We use the DDIM Sampler~\cite{song2020denoising} as the sampling strategy and sample 20 steps for both I-frame generation and the inpainting process of P-frames.
For optical flow estimation, we adopt the pre-trained model from the official GitHub repository of Flowformer~\cite{huang2022flowformer}.

In our experiments, we set the GoP size $g$ as 10. The weights of $\alpha$ and $\beta$ are set as 5 and 10, respectively. The $threshold$ for generating the inpainting mask is simply set as 0.5. We set the temperature $\tau$ as 20. In our inpainting mask generation module, we further apply the Gaussian Blur operation to expand the inpainting area of the occupancy map and use the min-pooling operation to generate the inpainting mask in latent space. 
All of our experimental results are generated on the machine with Tesla V100 GPU with 16GB GPU memory. We resize the input videos to $960\times540$ for evaluation.

\begin{figure*}[t]
  \centering
  \includegraphics[width=\linewidth]{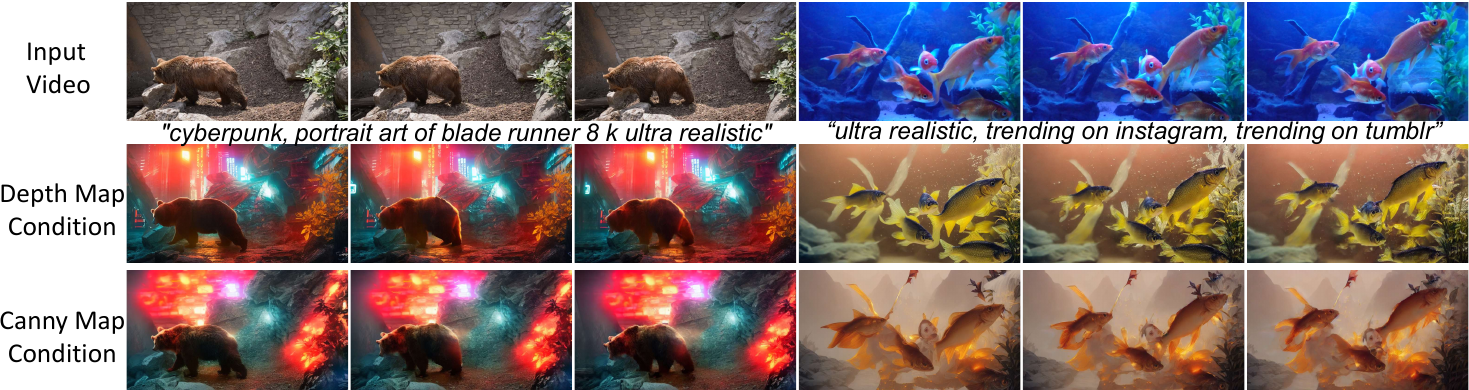}
      \caption{Generated results when using different conditions. The input videos are provided in the first row. The sentence below the input video is the input prompt of the StableDiffusion. The last two rows are the generated results, in which the middle row is the results when using the ControlNet with the depth map condition, and the last row contains the results when using the ControlNet with the canny map condition.}
  \label{fig:diffcond}
  \Description{diffcond.}
  \end{figure*}

\begin{figure}[t]
  \centering
  \includegraphics[width=\linewidth]{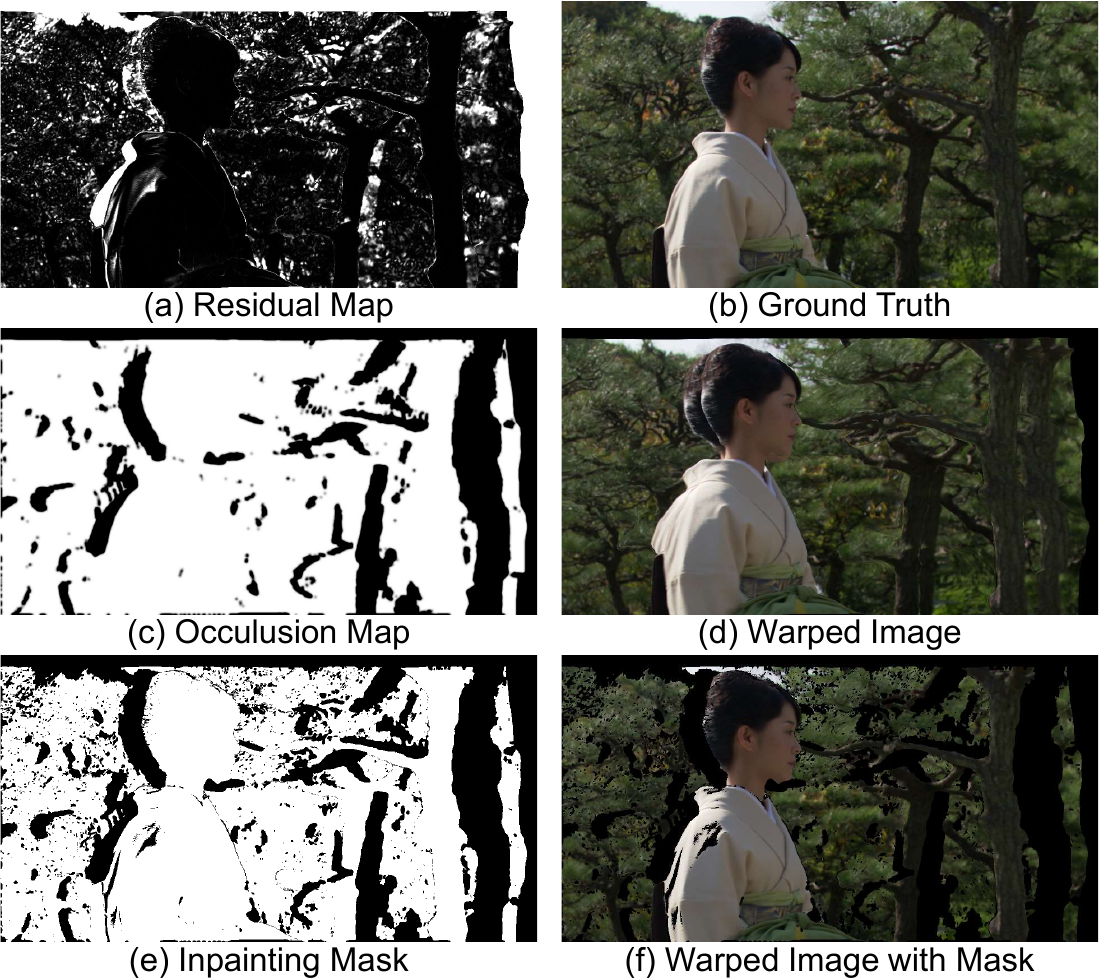}
      \caption{Visualization of our inpainting mask generation during our motion-guided P-frame generation (MgPG). (a) is the residual map that represents the difference between the warped frame and the ground truth frame. (b) is the ground truth of the current frame. (c) is the occlusion map calculated by using the forward warping operation. (d) is the wrapped frame by using the backward warping operation based on the reference frame. (e) is the inpainting mask calculated based on the residual map and the occlusion map. (f) is our warped frame with the inpainting mask.}
  \label{fig:inpaint}
  \Description{inpaint.}
  \end{figure}

\begin{table}[t!]
  \begin{center}
          \caption{User preference of Text2Video-Zero~\cite{khachatryan2023text2video}, CCPL~\cite{wu2022ccpl} and our proposed method.}\setlength{\tabcolsep}{5mm}{
  \begin{tabular}{|l|c|c|c|}
  \hline
                           & User Preference      \\
  \hline
     Text2Video-Zero       & 9.4\%  \\
  \hline
     CCPL                  & 15.8\%  \\
  \hline
     VideoControlNet (Ours) &  \textbf{74.7\%}      \\
  \hline
  \end{tabular}}
  \label{tab:userstudy}
  \end{center}
  \vspace{-4mm}
\end{table}

\subsection{Quantitative Results}

\textbf{User Study.}
We conduct a user study to evaluate the generation quality of our method, the SOTA diffusion-based video-to-video translation method Text2Video-Zero~\cite{khachatryan2023text2video} and the SOTA video style transfer method CCPL~\cite{wu2022ccpl}. Considering the previous non-diffusion-based video generation methods cannot support text-based instruction, we evaluate the SOTA video style transfer method CCPL~\cite{wu2022ccpl} by using the most relevant image of the text prompt as the style image.

We selected 100 video-prompt pairs for evaluation and use the official code and the default parameters for each method, in which the videos are from the DAVIS dataset~\cite{pont20172017}. For each user, 30 video-prompt combinations are randomly selected. Finally, we collected 720 votes from 24 users. The preference percentages of Text2Video-Zero~\cite{khachatryan2023text2video}, CCPL~\cite{wu2022ccpl} and our VideoControlNet are provided in Table~\ref{tab:userstudy}. Our VideoControlNet outperforms previous methods Text2Video-Zero~\cite{khachatryan2023text2video} and CCPL~\cite{wu2022ccpl} by a large margin and achieves 74.7\% user preference. It is observed that the generated videos of Text2Video-Zero lack temporal consistency. Although CCPL sometimes achieves comparable results on specific prompts, most generated videos are much worse than the results of our VideoControlNet.

\begin{table*}[t!]
  \begin{center}
          \caption{Quantitative results on the DAVIS dataset~\cite{pont20172017}.}\setlength{\tabcolsep}{3mm}{
  \begin{tabular}{|l|c|c|c|c|c|c|c|}
  \hline
                         &  FVD($\downarrow$) & IS($\uparrow$) & FID($\downarrow$) & CLIPSIM($\uparrow$) & LPIPS($\downarrow$) & Optical Flow Error($\downarrow$) & Speed ($\uparrow$)  \\
  \hline
     Text2Video-Zero     & 1670.39 & 13.23 & 119.01 & 25.66 & 0.56 & 17.99 & 0.19fps \\
  \hline
     Ours                & \textbf{981.99}  & \textbf{18.02} & \textbf{92.17}  & \textbf{26.14} & \textbf{0.50} & \textbf{7.91}  & \textbf{0.30fps} \\
  \hline
  \end{tabular}}
  \label{tab:object}
  \end{center}
  \vspace{-4mm}
\end{table*}

\vspace{1mm}

\noindent\textbf{Objective Metrics.}
We further provide the quantitative results on the DAVIS dataset~\cite{pont20172017} in Table~\ref{tab:object}, in which the video names (\textit{e.g.}, bus, dogs-jump, hike, paragliding-launch) are used as the text prompts. We use the objective metrics including Fr{\'e}chet Video Distance (FVD)~\cite{unterthiner2018towards}, Inception Score (IS), Fr{\'e}chet Inception Distance (FID)~\cite{heusel2017gans}, average CLIP~\cite{radford2021learning} similarity between video frames and text (CLIPSIM), LPIPS~\cite{zhang2018unreasonable} and the L2 distance between the optical flow ~\cite{ranjan2017optical} of the input video and the generated video (Optical Flow Error).

Considering the video style transfer method CCPL~\cite{wu2022ccpl} directly takes the video as input while our method and Text2Video-Zero~\cite{khachatryan2023text2video} only take the conditions (\textit{e.g.}, depth map, edge map) of the input video as input, we only compare our method with the SOTA video-to-video translation method Text2Video-Zero~\cite{khachatryan2023text2video}. We also report the running speed of both methods when generating the video with 40 frames and with the resolution of $960\times540$. 
It is observed that our method outperforms the SOTA diffusion-based method Text2Video-Zero~\cite{khachatryan2023text2video} in terms of all metrics including FVD, IS, FID, CLIPSIM, LPIPS and Optical Flow Error on the DAVIS dataset with faster inference speed, which demonstrate the effectiveness of our method.

\subsection{Qualitative Results}
We take the Text2LIVE~\cite{bar2022text2live} method as the baseline method to demonstrate the effectiveness of our proposed VideoControlNet. Text2Live~\cite{bar2022text2live} is recently proposed for text-guided video editing that adopts layered neural atlases~\cite{kasten2021layered}, which needs to fine-tune on each video and runs extremely slow. We use their official code and the provided configuration to generate the video.

The experimental results are provided in Figure~\ref{fig:results}, in which simple prompts below the input video are used for both Text2LIVE and our VideoControlNet. It is observed that the generated video of our method has better visual quality than the generation results from Text2LIVE due to the strong generation quality of StableDiffusion. For example, in the snow scene, the road of our generated video is more realistic than the output of Text2LIVE.  We also observe that the Text2LIVE method achieves good temporal smoothness due to its reliance on Layered Neural Atlases~\cite{kasten2021layered}. However, the generation quality of Text2Live outputs varies a lot on different types of videos. For example, in the nighttime scene, the road is illuminated without street lights.  Moreover, the inference speed of Text2Live is extremely slow and it even requires more than 10 hours for editing a single video, while our method generates the video at about 3.4 seconds per frame.
Finally, it is shown that the generated content of our method is consistent with the input video, which makes it able to use the corresponding optical flow to generate the output video by using our proposed VideoControlNet. The experimental results demonstrate that our method achieves better generation quality than the previous methods and can also keep the content consistent.

\subsection{Model Analysis}
\label{subsec:modelanalysis}

\noindent\textbf{Results when using Different Conditions.} 
In this work, we use the canny maps and depth maps as the condition information for the ControlNet~\cite{ni2023conditional} to generate the output video that has the same content as the input video. In Figure~\ref{fig:diffcond}, we provide the generated results when using different conditions. It is observed that when using the depth map as the condition of ControlNet, the generated results are more spatial and three-dimensional. For example, the generated fishes conditioned on the depth map are more three-dimensional than the results conditioned on the canny map. When using the canny map as the condition of ControlNet, the generated results contain more details (\textit{e.g.,} the bear fur). Therefore, we can condition the canny map for more detailed 2D image generation and use the depth map for generating 3D results.

\vspace{1mm}
\noindent\textbf{Generation of the Inpainting Mask.} 
Our inpainting masks are generated based on both occlusion maps and residual maps for generating the newly occurred areas for the current frame. To better illustrate our inpainting mask generation process, we visualize the residual map, the occlusion map and our final inpainting mask $I_i$ in Figure~\ref{fig:inpaint}. It is observed that most occlusion areas can be found in our occlusion map (\textit{e.g.,} the left side of the woman), which is generated by using the forward warping operation based on the reverse optical flow. However, the motion information between the neighbor P frames may be very large and thus the optical flow estimation network may not generate accurate motion information. To this end, we additionally use the residual map for detecting the occlusion areas. Additionally, the residual map shown in Figure~\ref{fig:inpaint} is very sparse. Therefore, it is also not reliable if we only use the residual map to generate the inpainting mask. To this end, we use both the residual map and the occlusion map for generating the inpainting mask. As shown in Figure~\ref{fig:inpaint}(f), the occlusion areas are well masked by the inpainting mask, which demonstrates the effectiveness of our inpainting mask generation module.

\begin{table}[t!]
  \begin{center}
          \caption{Running Time of different modules in which we use 20 sampling steps for the StableDiffusion with ControlNet. ``Pixel to Latent" denotes encoding the image to latent space for the diffusion networks. ``Latent to Pixel" denotes decoding the image from latent space. We also provide the inference time of I-frame generation, P-frame generation and B-frame generation. The average time is calculated when the GoP size is set as 10.}\setlength{\tabcolsep}{3mm}{
  \begin{tabular}{|l|c|}
  \hline
                                & Time      \\
  \hline
    StableDiffusion with ControlNet   & 13.7s     \\
  \hline
    Pixel to Latent                & 0.13s     \\
  \hline
    Latent to Pixel                & 0.01s     \\
  \hline
    Inpainting Mask Generation  & 0.97s     \\
  \hline
    I-frame Generation          & 13.7s     \\
  \hline
    Motion-Guided B-frame Interpolation          & 1.9s      \\
  \hline
    Motion-Guided P-frame Generation          & 14.8s     \\
  \hline
    Average Time Per Frame      & 3.4s     \\
  \hline
  \end{tabular}}
  \label{tab:time}
  \end{center}
  \vspace{-4mm}
\end{table}

\vspace{1mm}
\noindent\textbf{Running Speed.} 
The detailed running time of different modules is provided in Table~\ref{tab:time}.
We evaluate our inference speed on the machine with a single Tesla V100 GPU. The input videos are resized to the resolution of $960 \times 540$. The GoP size $g$ is set as 10 and we use 20 diffusion steps. It is observed that the StableDiffusion model with ControlNet needs 13.7s for generating the I-frame or inpainting the occlusion areas of P-frames. The inpainting mask generation module costs 0.97s, in which the main part of the time is spent on the optical flow estimation network~\cite{huang2022flowformer}. Due to the fast speed of our motion-guided B-frame interpolation which costs 1.9s per frame, our average generation speed is about 3.4s per frame when generating 4 GoP of images with $g=10$. Therefore, our proposed VideoControlNet framework is even faster than generating the output video frame-by-frame by using StableDiffusion Model with ControlNet, which requires 13.7s per frame.

\vspace{1mm}
\noindent\textbf{Discussion of Applications.} 
Our VideoControlNet achieves video-to-video translation by using motion information, in which the content of the input video should be consistent with the output video. Therefore, the most straightforward application is the video style transfer, which can achieve various styles based on different given prompts. Additionally, our method can also achieve video editing with the extra mask of the object to be edited. In conclusion, our method can be regarded as a video version of the ControlNet, that is able to control the content and motion information based on the given input video and prompts.  Unfortunately, due to the detailed motion information, the condition should be more strong that can control the details of the output photo. Therefore, the segmentation map or human pose can not be used as the condition of our VideoControlNet. More results are provided in our appendix.

\section{Conclusion}

In this work, we propose a new video-to-video translation framework called VideoControlNet based on the StableDiffusion model with ControlNet, in which the motion information is adopted for better content consistency. We first generate the I-frame and then divide the rest frames into different groups of pictures (GoP), in which the last frame of each GoP is set as the P-frame and others are B-frames. For the P-frame generation, we propose the motion-guided P-frame generation (MgPG) module to prevent the regeneration of the occlusion areas and only inpaint the occlusion areas. For the B-frame generation, we propose the motion-guided B-frame interpolation (MgBI) module to directly interpolate the B-frames by using the two nearest I/P frames as the reference frame. The experimental results demonstrate that our VideoControlNet framework achieves impressive video generation results with high-quality content and good content consistency with the motion information from the input video. In our future work, we will study adding more learnable networks for better content consistency.


\bibliographystyle{ACM-Reference-Format}
\bibliography{sample-base}

\appendix








\begin{figure*}[t!]
  \centering    
  \includegraphics[width=\textwidth]{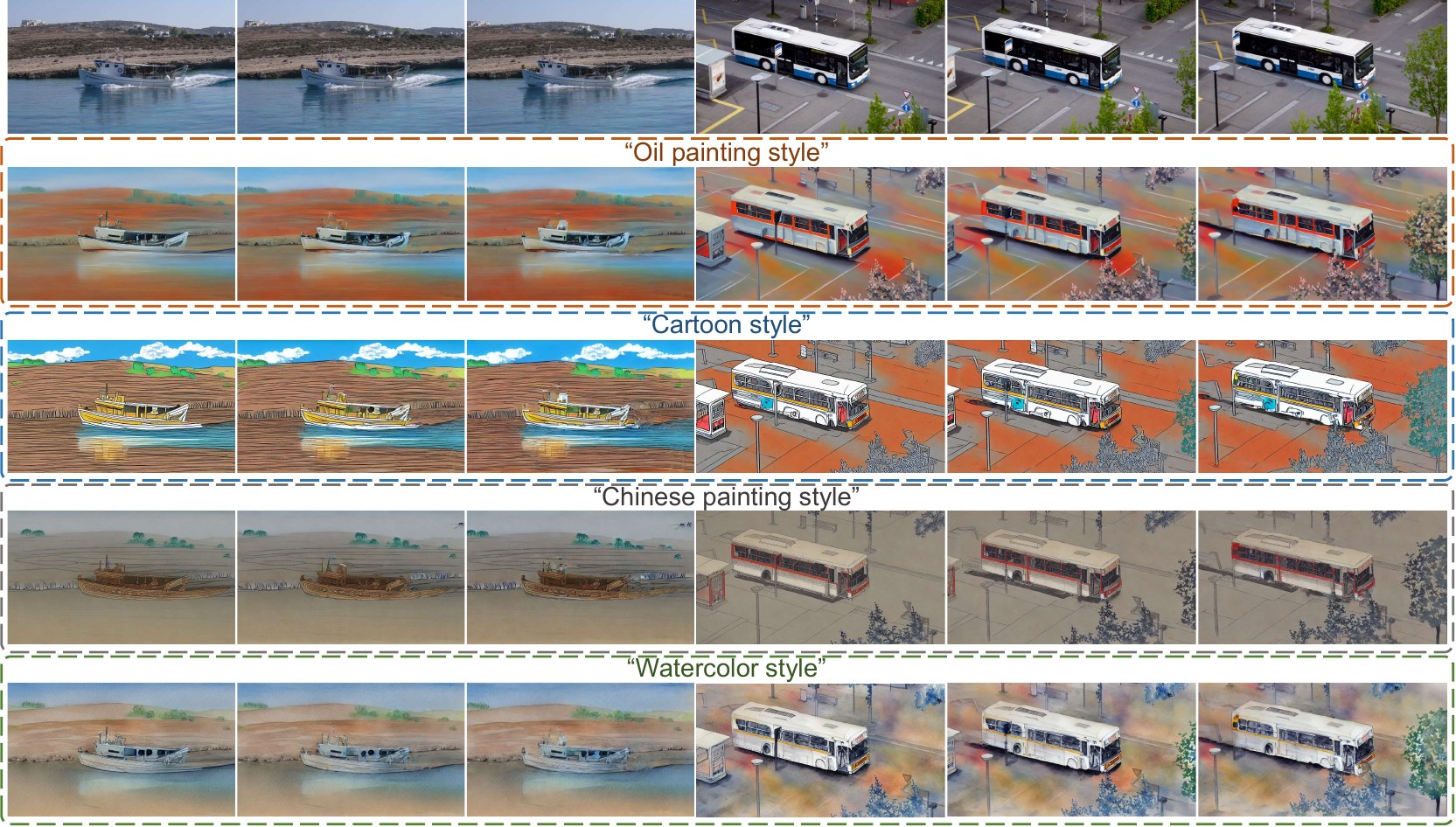}
  \caption{Video style transfer results by using the prompts to control the style. The first row is the input video and the other rows are the generated videos based on the input video and given prompts.}
  \Description{x.}
  \label{fig:styletransfer}
\end{figure*}

\setcounter{section}{0}

\renewcommand\thesection{\Alph{section}}

\section{Video Examples}
In our main paper, we provide some examples of the generated videos in Figure~1, in which only a few images in the video are shown. To further demonstrate the effectiveness of our proposed VideoControlNet, we provide the video examples in our \href{https://vcg-aigc.github.io/}{\textcolor{magenta}{project page}}. It is observed that our generated videos have the same motion and content as the input video and our VideoControlNet can generate various types of videos based on any type of input video and prompts.

\begin{figure}
  \centering    
  \includegraphics[width=\linewidth]{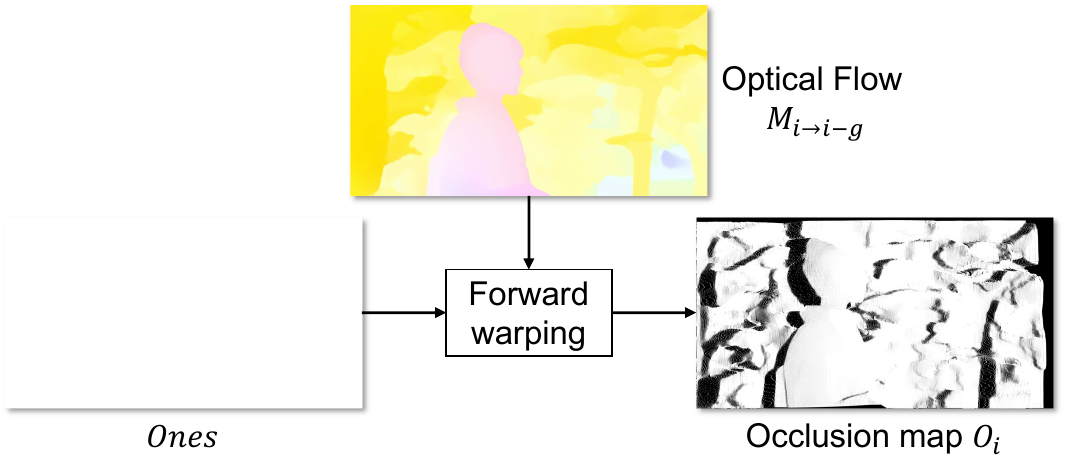}
  \caption{Illustration of our occlusion map generation by using forward warping operation. Taking a map full of values one and the optical flow $M_{i \rightarrow i-g}$ as input, the forward warping operation generates the occlusion map $O_i$.}
  \Description{x.}
  \label{fig:forwardwarp}
\end{figure}

\begin{figure*}[t!]
  \centering    
  \includegraphics[width=\linewidth]{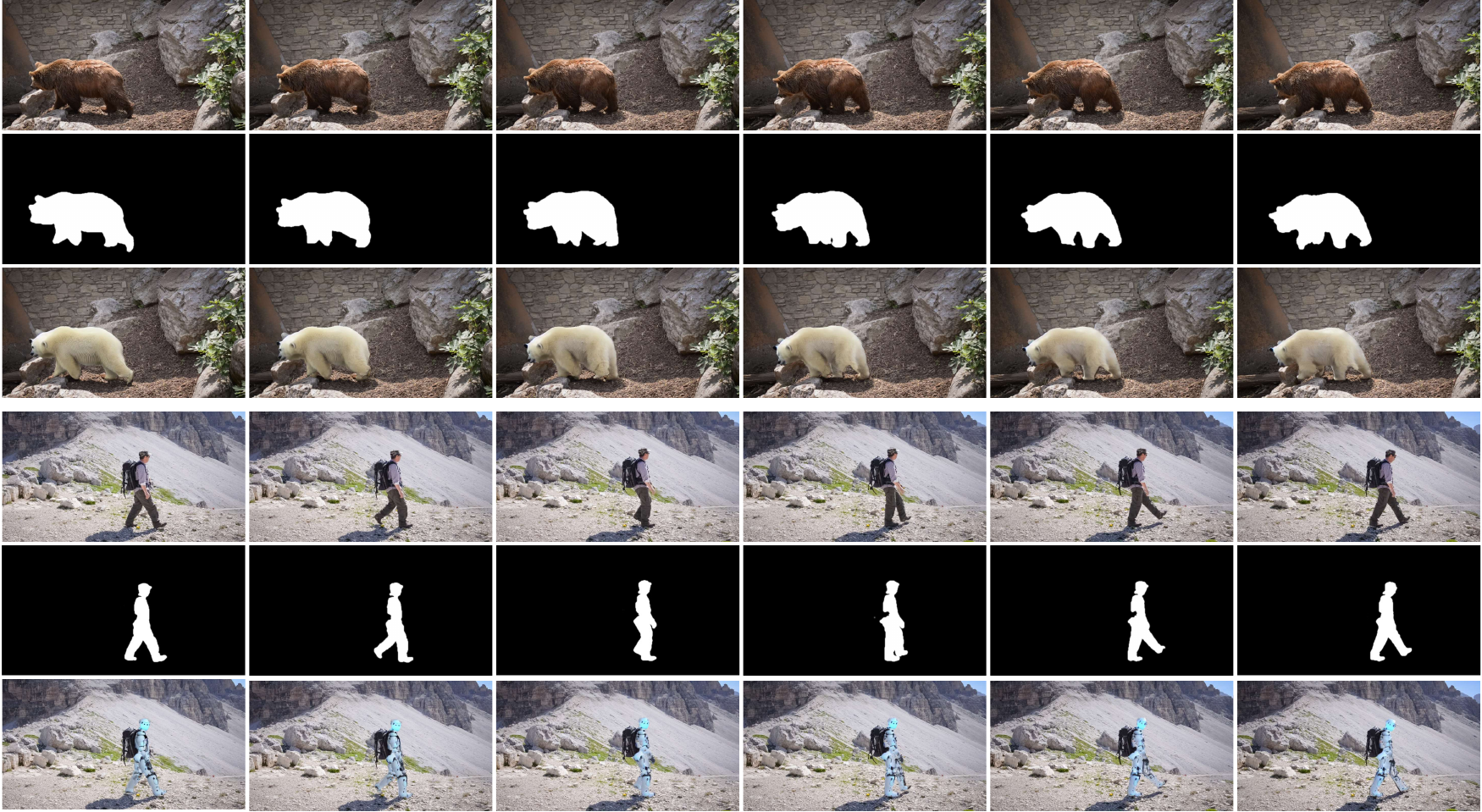}
  \caption{The video editing results of our proposed method. The first row and the fourth row are the input videos. The second row and the fifth row are the masks to be edited. The third row and the last row are the output videos, which are generated by using the prompts ``A robot is hiking" and ``polar bear", respectively.}
  \Description{x.}
  \label{fig:edit}
\end{figure*}

\section{Illustration of Occlusion Map Generation}
In our inpainting mask generation module, a $Ones$ map is forward warped based on the optical flow $M_{i \rightarrow i-g}$ for generating the occlusion map $O_i$. Backward warping is the widely used operation for motion compensation and the optical flows are always generated for backward warping.  However, we cannot figure out the locations that are newly occurred in the current frame by using the backward warping operation. Therefore, we use the forward warping operation to figure out the newly occurred areas. As shown in Figure~\ref{fig:forwardwarp}, the background is moving towards the left side and the woman has only little movement. Therefore, by using the forward warping based on the $Ones$ map, the values on the left side of the woman move to the left, and the values inside the woman still stay inside, which makes it easy to find out the occlusion areas at the left side of the woman. By using both the occlusion map and the residual map, we can find out the areas to be inpainted.

\section{Applications}
As discussed in our main paper, our VideoControlNet framework is able to achieve applications like style transfer and video editing. Therefore, we provide more visualization results in the supplementary materials.

\subsection{Video Style Transfer}
As shown in Figure~\ref{fig:styletransfer}, we provide the style transfer results of our proposed VideoControlNet. The first row contains the input video and the other rows are the generated results of different styles, in which the styles are controlled by the prompts. It is observed that our VideoControlNet framework is able to translate the input video into different styles including the oil painting style, cartoon style, Chinese painting style and watercolor style, which further demonstrate the effectiveness of our method. We also provide more video style transfer results in our project.

\subsection{Video Editing}
The video editing results are provided in Figure~\ref{fig:edit}. Given the input video and the masks that needed to be edited, our VideoControlNet can generate the output video that edits the content based on the given masks and prompts. For example, for the output video in the last row of Figure~\ref{fig:edit}, we masked the man that is hiking and the corresponding prompt is ``a robot is hiking". Therefore, our method can generate a robot that is hiking. Note that our masks are generated by using the official code of ``Segment Anything"~\cite{kirillov2023segment}. We also provide more video editing results in our project.

\end{document}